\documentclass[]{spie}  

\usepackage[]{graphicx}

\title{Examining Performance of Sketch-to-Image Translation Models with Multiclass Automatically Generated Paired Training Data} 

\author{Dichao Hu
\skiplinehalf
College of Computing, Georgia Institute of Technology, 801 Atlantic Dr NW, Atlanta, GA 30332, USA \\
}

\authorinfo{Further author information: (Send correspondence to Dichao Hu)\\Dichao Hu: E-mail: hudichao1998@gatech.edu, Telephone: 65 8266 7534\\}

\begin{document} 
  \maketitle 

\begin{abstract}
Image translation is a computer vision task that involves translating one representation of the scene into another. Various approaches have been proposed and achieved highly desirable results. Nevertheless, its accomplishment requires abundant paired training data which are expensive to acquire. Therefore, models for translation are usually trained on a set of paired training data which are carefully and laboriously designed. Our work is focused on learning through automatically generated paired data. We propose a method to generate fake sketches from images using an adversarial network and then pair the images with corresponding fake sketches to form large-scale multi-class paired training data for training a sketch-to-image translation model. Our model is an encoder-decoder architecture where the encoder generates fake sketches from images and the decoder performs sketch-to-image translation. Qualitative results show that the encoder can be used for generating large-scale multi-class paired data under low supervision. Our current dataset now contains 61255 image and (fake) sketch pairs from 256 different categories. These figures can be greatly increased in the future thanks to our weak reliance on manually labelled data.
\end{abstract}




\section{Introduction}
Image-to-image translation is defined as "the task of translating one possible representation of a scene into another, given sufficient data." \cite{pix2pix2017} Although various methods have been proposed and achieved impressive results, most of them rely on a large number of paired training data \cite{xie15hed,fine-grained,Tylecek13,Cordts2016Cityscapes,
zhu2016generative,Laffont14} to reach high performance, while the acquisition of such paired data is costly. Therefore it is highly desirable if we can efficiently obtain a large number of paired training data without much human effort. In this paper, we explore the application of generative adversarial networks in automatic acquisition of paired data for sketch-to-image translation. Previous successful translation models include (but are not limited to) pix2pix \cite{pix2pix2017} for paired translation and cycleGAN \cite{CycleGAN2017} for unpaired translation. Pix2pix is a conditional adversarial network that receives an image from the source domain as conditional signal and generates a corresponding image of the target domain. However, its success relies heavily on abundant paired training data. On the other hand, cycleGAN can be used for unpaired translation, but its performance drops significantly on difficult translation tasks that involve radical changes in content of images. Our model is a combination of the two, aiming to let them complement each other. We use cycleGAN as an autoencoder \cite{hinton2006reducing} and specify its input as the original image(s) and output as the generated sketch(s). Following Zhu et al.\cite{CycleGAN2017}, our goal is to learn two set-based mappings $G : X\rightarrow Y$ and $F : Y\rightarrow X$ where $X$ is the set of images and $Y$ is the set of sketches. Therefore, $G \circ F$ forms an autoencoder that takes the images as input, encodes the images into sketches by passing through the generator $G$, and then decodes sketches back to images by passing through another generator $F$. In this case, the encoded representation has the same dimension as the input, but with a much sparser density distribution as all values are pushed towards either 0 (black) or 255 (white). Once we have the autoencoder trained, we use the generator $G$ to generate approximate sketches from input images and then use (real) image and (fake) sketch pairs as our training data for another network, the pix2pix net. 
Such architecture design is motivated by our observation that image-to-sketch (I2S) translation is an easier task compared to sketch-to-image (S2I) translation. Our results show that the cycleGAN encoder can perform I2S translation even if with a small (500 images) single-class training set and capable of generalizing to large multi-class (60000+ images of 256 different categories) test set. In this way, we can access numerous automatically generated paired training data without human labelling efforts. These generated paired data are then used for training the pix2pix decoder.
\label{headings}

\section{Related Work}

\subsection{Cycle-consistent Adversarial Network}
The cycle-consistent adversarial network architecture (cycleGAN) was recently proposed by Zhu et al.\cite{CycleGAN2017} for unpaired image-to-image translation. A pair of mirror adversarial networks $F$ and $G$ are constructed for  translation from one domain to the other and vice versa. In addition to the adversarial loss  \cite{goodfellow2014generative} that constrains the generated images to be indistinguishable from real images, a key feature is its cycle-consistency loss forcing the double-translated image $F(G(x))$ to be close to the original image. In this paper, we explore usage of cycleGAN as an autoencoder to learn a class-invariant encoding of images from 256 different categories. The learned encoder trained under low supervision is then responsible for generating sketches of larger scale and multiple categories.

\subsection{Adversarial Network with Conditional Signals}

Application of adversarial networks under the conditional settings has been expored in previous works \cite{mirza2014conditional,zhu2017toward,pix2pix2017}. The motivation is that results can be better controlled based on the conditional signal passed into the adversarial network as additional features. In Mirza et al. \cite{mirza2014conditional}, an adversarial network was trained on MNIST images, conditioned on their class labels represented as one hot vectors. In Isola et al.\cite{pix2pix2017}, similar idea was applied to the image translation domain. Training data were represented pairwise where the adversarial network (pix2pix) generates images from the target domain conditioned on their counterparts in the source domain. Similar approaches\cite{zhu2017toward} include combining a translation network with a variational autoencoder \cite{kingma2013auto} to increase in-class variability of generated images. In our work, we represent as the conditional signal the combination of labels and images. Unlike in Isola et al.\cite{pix2pix2017}, however, we replace the manually created paired sketch-and-image training data with our paired fake-sketch-and-real-image training data where fake sketches are generated by the cycleGAN encoder.

\section{Formulation}
\begin{figure}[htbp]
\centering
\includegraphics[width = 0.8\textwidth]{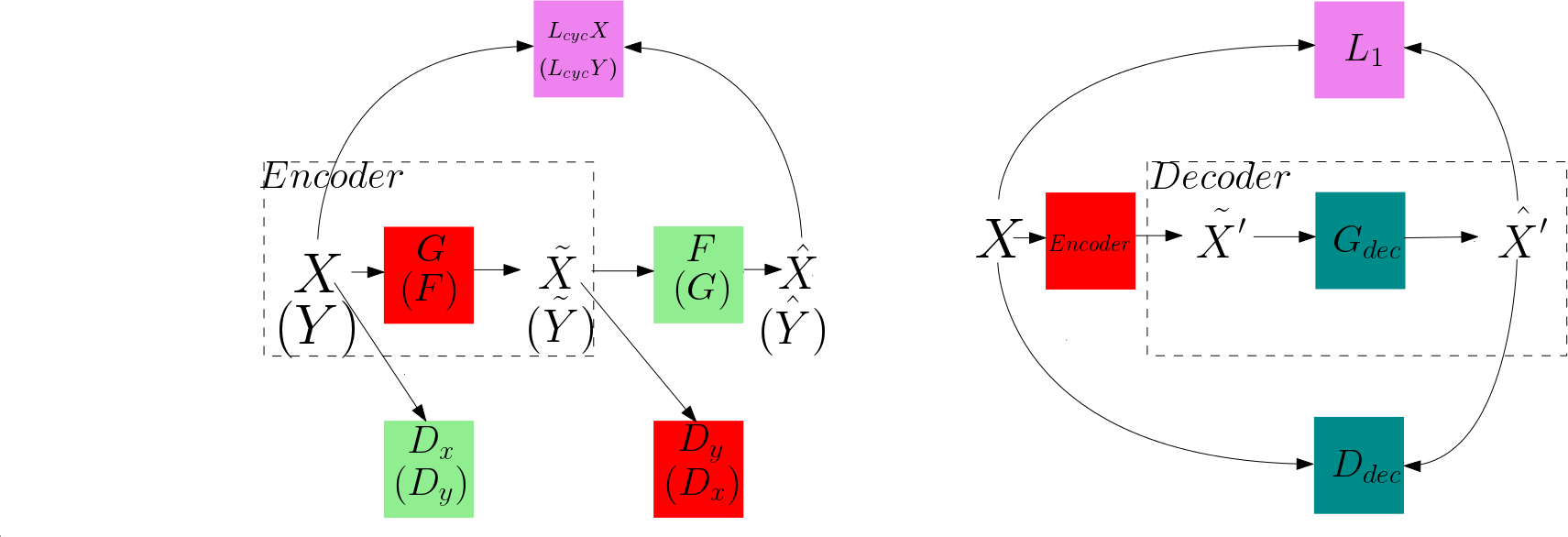}
\caption{An illustration of our encoder-decoder architecture. The image on the left indicates the cycleGAN training procedure. A pair of GANs are trained together. $G$ and $D_{y}$ are generator and discriminator for the forward (image-to-sketch) direction, while $F$ and $D_{x}$ denote the backward direction. $L_{cyc}$ is the penalty term to maintain cycle-consistency. $\displaystyle X,\tilde{X},\hat{X}$ stand for the original image, generated sketch and double-translated image, respectively. After training is finished, only $G$ is kept as $Encoder$ and other parts are discarded. In the right image, the learned $Encoder (G)$ is then used for generating fake sketches for training the pix2pix $Decoder (G_{dec})$.}
\label{fig:model}
\end{figure}
Our model is an encoder-decoder architecture and can be further broken down into the two adversarial networks aforementioned, cycleGAN for encoding and pix2pix for decoding. For simplicity, we will use \textit{Encoder} and \textit{Decoder} for the cycleGAN and pix2pix, respectively. \textit{Encoder} takes an input image and transforms it into its sketch counterpart. \textit{Decoder} takes both the generated sketch and the class label of the image as input and its goal is to reconstruct the original image. Full training process can be represented as a two-step pipeline (Figure~\ref{fig:model}). The first step (\textit{Encoder}) is to learn two mappings $G: X \rightarrow Y$ and $F : Y \rightarrow X$ where X and Y corresponds to a set of images and a set of sketches respectively. After training, $G$ is used as \textit{Encoder}, and $F$ is discarded. Note that here images and sketches are not paired.  Then the second step (\textit{Decoder}) is to learn another mapping $H : \{G(X),l(X)\} \rightarrow X$ where $G(X)$ is the set of generated sketches and $l(X)$ is the set of labels.
\subsection{Objective for \textit{Encoder}}
The full objective for \textit{Encoder} exactly follows  the original paper\cite{CycleGAN2017}:\\
$L(G;F;D_X;D_Y)$\\ $=L_{GAN}(G; D_Y ; X; Y )
+ L_{GAN}(F; D_X; Y; X);
+ \lambda \cdot Lcyc(G; F );$\\
where \\$L_{GAN}(G;D_Y;X;Y)$ \\$=E_{y} [log D_Y (y)]
+E_{x} [log(1 - D_Y (G(x))]$
\\$L_{cyc}(G;F) $\\$=E_{x} [||F (G(x)) - x||_1]
+E_{y} [||G(F (y)) - y||_1];$\\
Our aim is to solve:\\
$G^\ast, F^\ast = argmin_{G;F}max_{D_X;D_Y}L(G;F;D_X;D_Y).$\\
Here $D_X$ and $D_Y$ are two discriminators for images and sketches, respectively. $L_{cyc}$ stands for the cyclic penalty that constrains the double-translated representation to be close to the original, in terms of $L_1$ distance. The goal is to learn a universal image-to-sketch encoding $G$, which is then used as \textit{Encoder} after training is done.
\subsection{Objective for \textit{Decoder}}
During training, we find that using least square loss \cite{mao2017least} produces more robust results. The modified objective can be expressed as\\ 
$G^\ast = argmin_{G}max_{D} L_{cGAN}(G;D) + \lambda L_{L1}(G);$\\
where\\ 
$ \tilde{y} = G(x;l(x));$\\
$L_{L1}(G) = E_{x;y;l(x)}[||y - \tilde{y}||_1];$\\$
L_{cGAN}(G; D)$ \\$=E_{x;y;l(x)}[(D(x;l(x);y))^2]+E_{x;l(x)}[(1-D(x; l(x);\tilde{y}))^2];$\\
Here $D(x;l(x);y)$ means that we pass both the label $x$ and sketch $l(x)$ signal into the discriminator $D$. We use $ \tilde{y}$ to represent generated sketches. The adversarial loss \cite{goodfellow2014generative} encourages to generate realistic images from sketches. We use $L_1$ penalty for both \textit{Decoder} and \textit{Encoder} to encourage less blurring.

\section{Implementation}
It is hard to present every part of our model in detail because it is constructed on top of deep convolutional neural networks with multiple skip connections. Instead, we provide a series of abbreviated layer notation similar to Isola et al.\cite{pix2pix2017} to summarize the whole architecture. U, D, R stands for upsampling layer(stride = 2), downsampling layer (stride = 2) and residual blocks, respectively.
\subsection{\textit{Encoder} Architecture}
We use the same architecture for \textit{Encoder} as in Zhu et al. \cite{CycleGAN2017}, where details can be found. \\
\textbf{Generator}\\
D32-D64-D128-D256-D256-D256-D256-D256-U512-U512-U512-U512-U256-U128-U64-U3\\
\textbf{Discriminator}\\
D64-D128-D256-D512\\
Both inputs (images) and outputs (sketches) of the generator are of size 256-by-256 in gray scale. The discriminator takes a 256-by-256 sketch (either real or fake) and outputs a real-value as the classification score. We use a batch size of 4, regularization strength $\lambda _{cyc}$ of 10, and initial learning rate of $10^{-4}$. We repeatedly decrease the learning rate to $\frac{1}{10}$ when we approximately achieve convergence using the current learning rate. In addition, we use Relu activation \cite{nair2010rectified} for discriminator, and leaky Relu ($\alpha$ = 0.2) for generator. We apply dropout rate of 0.5, instance normalization \cite{ulyanov1607instance}, and U-net concatenation \cite{ronneberger2015u} as in the original paper \cite{CycleGAN2017}.

\subsection{\textit{Decoder} Architecture}
We modify the \textit{Decoder} architecture by adding residual blocks \cite{he2016deep} to accommodate the increase in data diversity.\\
\textbf{Generator}\\
D64-D128-D256-D512-D512-D512-D512-D512-
\\
U1024-R1024-R1024-U1024-R1024-R1024-U1024-U1024-U512-U256-U128-U3\\
\textbf{Discriminator}\\
D64-D128-D256-D512\\
We use a batch size of 4, initial learning rate of $10^{-6}$ and regularization strength $\lambda$ of 100. Learning rate is fine-tuned the same way as in the  \textit{Encoder}. Also, instead of passing the sketch alone as input to the generator, we broadcast the one-hot label representation and concatenate it with the sketch. Therefore, the input to both the generator and the discriminator is a 256-by-256-by-4 tensor constructed from the 256-by-256-by-3 sketch concatenated with its class label (256-by-256-by-1 after broadcasting).

\subsection{Dataset and Details}
\begin{figure}[htbp]
\centering
\includegraphics[width = 0.5\textwidth]{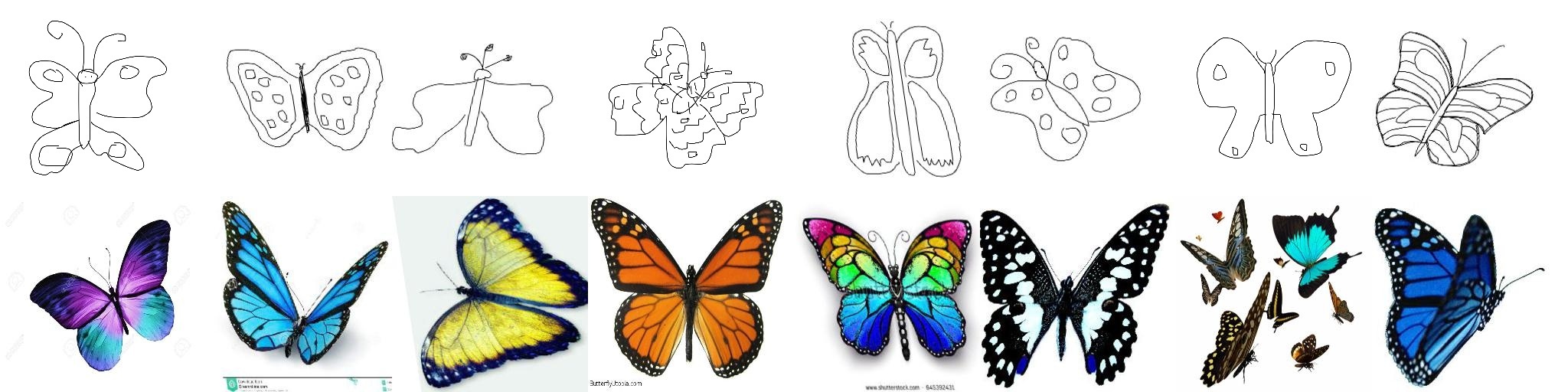}
\caption{We use unpaired images and sketches of butterflies for training the cycleGAN \textit{Encoder}. The first row are sketches of butterflies. The second row are images of butterflies.}
\label{fig:training_data}
\end{figure}

We discuss training details for \textit{Encoder} and \textit{Decoder} separately. Our goal for \textit{Encoder} is to learn a mapping $G: X\rightarrow Y$ where $X$ is the set of images and $Y$ is the set of sketches. Translation is therefore equivalent to removing unnecessary details in the original image while preserving its high-level structure so that we can reconstruct the original image from generated sketch. During training we observe that \textit{Encoder} performance depends heavily on color variability of images. Thus our dataset for training \textit{Encoder} solely consists of 500 images and 100 sketches\cite{eitz2012hdhso} of the butterfly category (Figure~\ref{fig:training_data}). During testing, we apply \textit{Encoder} to all images of 256 categories. Results will be demonstrated in the next section.\\
Unlike \textit{Encoder}, training data for \textit{Decoder} are of much larger scale and higher diversity. We construct our dataset directly from Google Image with approximately 250 images for each of the 256 categories. Images are almost evenly distributed across different categories.

\section{Results}
\begin{figure}[htbp]
\centering
\includegraphics[width = 0.5\textwidth]{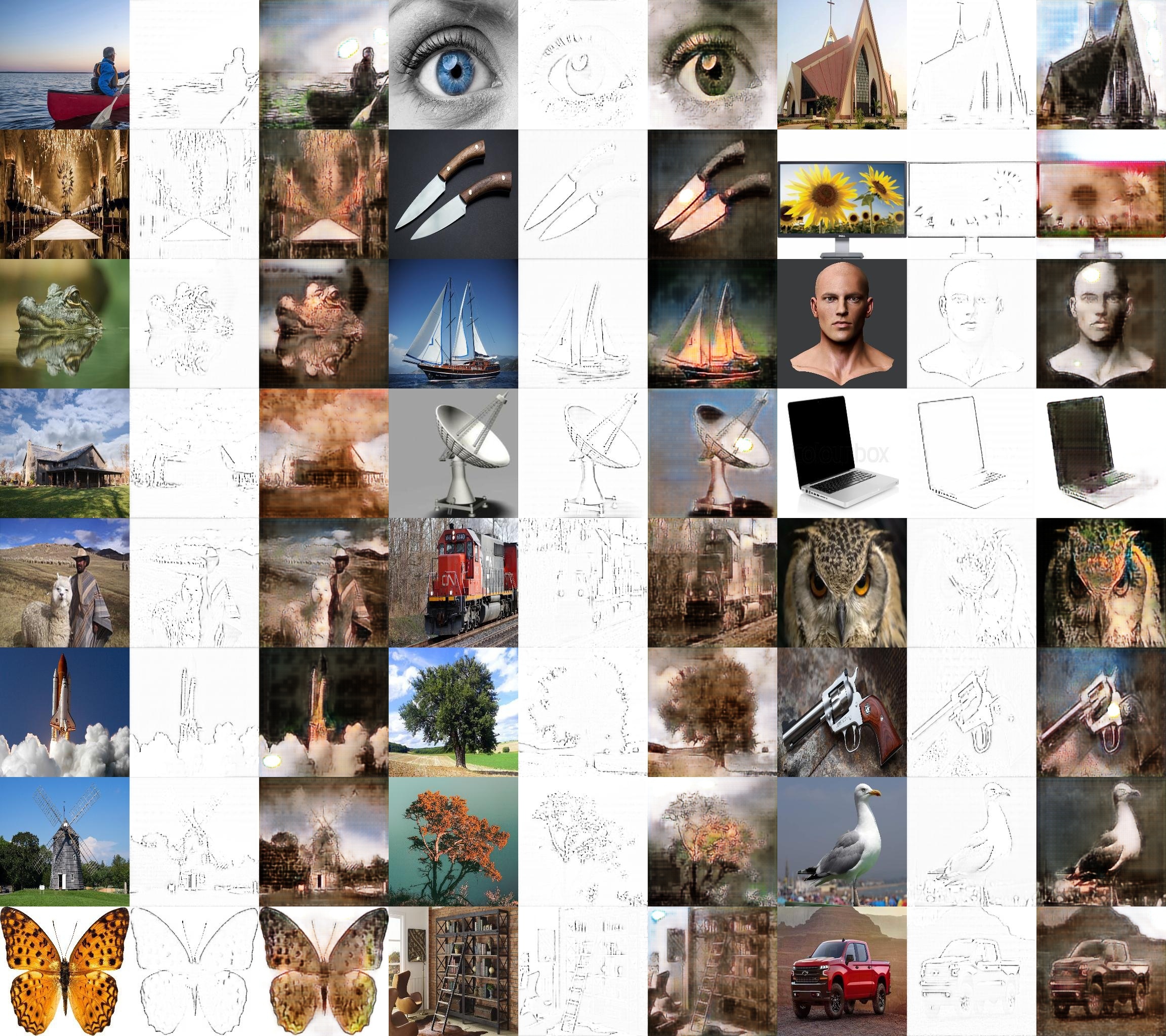}
\caption{Results for both the cycleGAN and pix2pix generated using test set images from 256 categories. For every three consecutive columns, the left column contains input images for cycleGAN, the middle column contains output sketches from cycleGAN which are also fed into the pix2pix net, and the right column contains output images from pix2pix.}
\label{fig:results_all}
\end{figure}
Here we present our results for both \textit{Encoder} and \textit{Decoder} in Figure~\ref{fig:results_all}. The following section will be focused on qualitative analysis on results we have achieved.

\subsection{\textit{Decoder}}
The pix2pix network has already been widely used in translation and achieved highly desirable results in different domains. As we have expected, the generated images still pertain high fidelity despite the increase in diversity of training images. We also find that \textit{Decoder} is capable of handling corrupted input due to  the undesired behaviour of \textit{Encoder}. However, one obvious aspect of performance which goes down significantly with the data increase is the color diversity of the reconstructed image. The \textit{Decoder} learns to be very careful on usage of different colors to make its generated images look realistic. \textit{Decoder} also performs poorly with unfamiliar data, such as arbitrarily hand-drawn sketches. 

\subsection{\textit{Encoder}}
However, misbehaviour of \textit{Decoder} is not mainly caused by the pix2pix network itself, but by the misbehaviour of the cycleGAN \textit{Encoder}. As discussed in 4.3, \textit{Encoder} is trained only on a specific class of images (butterflies in our work), and can easily be generalized to other categories. For example, it can translate an image of a male even if it has never observed any during training. Nevertheless, behaviour of \textit{Encoder} is highly susceptible to shading effects. Unexpected local shading in images leads to failure in local translation. 

\subsection{Fake Sketch Versus Real Sketch}
One thing we haven't discussed is the usage of these fake sketches generated by \textit{Encoder}. A natural solution is to take use of transfer learning techniques to help the translation using real sketches. Specifically, we can represent the set of fake sketches as from the source domain where pair labels are automatically generated  and sufficient in quantity, and the set of real sketches as from the target domain where paired data are scarce. Since our work is mainly focused on generation of paired data, we only explored zero-shot transfer and will leave the rest for future research. Namely, we trained our model only on fake sketches and then directly applied the model to real sketches. Such a naive approach didn't produce a good result on real sketches due to the difference in distribution of real and fake sketches. We can demonstrate this by comparing the distribution of pixel values in fake sketches with that in real sketches. Results show that about 90 percent of pixel values in real sketches are either absolutely white (255) or black (0), but only about 30 percent in fake sketches. Most pixel values lie in between this range. A slightly better approach is to add a preprocessing step before passing the encoded sketch to \textit{Decoder} by using thresholding and therefore force all pixel values to be binary (either 0 or 255 depending on the threshold). However, preprocessed sketches are still distinguishable from real sketches since the later can be constructed in arbitrarily distorted shape. Nevertheless, we strongly assume that our automatically-generated fake sketches are potentially helpful in real sketch-to-image translation, if appropriate transfer techniques are applied.
\section{Conclusion}
In this paper, we proposed an encoder-decoder architecture that first generates multi-class paired data and then performs sketch-to-image translation. Desirable and robust results for both generated sketches and images can be obtained under low supervision. However, performance drops significantly when the model trained on fake sketches is applied directly to real sketches, due to their difference in distribution. Our future research will be focused on applying appropriate transfer learning techniques to fill this gap.

\bibliographystyle{spiebib}

\begin{thebibliography}{10}

\bibitem{pix2pix2017}
Isola, P., Zhu, J.-Y., Zhou, T., and Efros, A.~A., ``Image-to-image translation
  with conditional adversarial networks,'' {\em CVPR}  (2017).

\bibitem{xie15hed}
"Xie, S. and Tu, Z., ``Holistically-nested edge detection,'' in [{\em
  Proceedings of IEEE International Conference on Computer
  Vision}{\nolinebreak\hspace{0.1em}]},  (2015).

\bibitem{fine-grained}
Yu, A. and Grauman, K., ``{F}ine-{G}rained {V}isual {C}omparisons with {L}ocal
  {L}earning,'' in [{\em Computer Vision and Pattern Recognition
  (CVPR)}{\nolinebreak\hspace{0.1em}]},  (June 2014).

\bibitem{Tylecek13}
Radim~Tyle{\v c}ek, R.~{\v S}., ``Spatial pattern templates for recognition of
  objects with regular structure,'' in [{\em Proc.
  GCPR}{\nolinebreak\hspace{0.1em}]},  (2013).

\bibitem{Cordts2016Cityscapes}
Cordts, M., Omran, M., Ramos, S., Rehfeld, T., Enzweiler, M., Benenson, R.,
  Franke, U., Roth, S., and Schiele, B., ``The cityscapes dataset for semantic
  urban scene understanding,'' in [{\em Proc. of the IEEE Conference on
  Computer Vision and Pattern Recognition (CVPR)}{\nolinebreak\hspace{0.1em}]},
   (2016).

\bibitem{zhu2016generative}
Zhu, J.-Y., Kr{\"a}henb{\"u}hl, P., Shechtman, E., and Efros, A.~A.,
  ``Generative visual manipulation on the natural image manifold,'' in [{\em
  Proceedings of European Conference on Computer Vision
  (ECCV)}{\nolinebreak\hspace{0.1em}]},  (2016).

\bibitem{Laffont14}
Laffont, P.-Y., Ren, Z., Tao, X., Qian, C., and Hays, J., ``Transient
  attributes for high-level understanding and editing of outdoor scenes,'' {\em
  ACM Transactions on Graphics (proceedings of SIGGRAPH)}~{\bf 33}(4) (2014).

\bibitem{CycleGAN2017}
Zhu, J.-Y., Park, T., Isola, P., and Efros, A.~A., ``Unpaired image-to-image
  translation using cycle-consistent adversarial networks,'' in [{\em Computer
  Vision (ICCV), 2017 IEEE International Conference
  on}{\nolinebreak\hspace{0.1em}]},  (2017).

\bibitem{hinton2006reducing}
Hinton, G.~E. and Salakhutdinov, R.~R., ``Reducing the dimensionality of data
  with neural networks,'' {\em science}~{\bf 313}(5786),  504--507 (2006).

\bibitem{goodfellow2014generative}
Goodfellow, I., Pouget-Abadie, J., Mirza, M., Xu, B., Warde-Farley, D., Ozair,
  S., Courville, A., and Bengio, Y., ``Generative adversarial nets,'' in [{\em
  Advances in neural information processing
  systems}{\nolinebreak\hspace{0.1em}]},   2672--2680 (2014).

\bibitem{mirza2014conditional}
Mirza, M. and Osindero, S., ``Conditional generative adversarial nets,'' {\em
  arXiv preprint arXiv:1411.1784}  (2014).

\bibitem{zhu2017toward}
Zhu, J.-Y., Zhang, R., Pathak, D., Darrell, T., Efros, A.~A., Wang, O., and
  Shechtman, E., ``Toward multimodal image-to-image translation,'' in [{\em
  Advances in Neural Information Processing
  Systems}{\nolinebreak\hspace{0.1em}]},  (2017).

\bibitem{kingma2013auto}
Kingma, D.~P. and Welling, M., ``Auto-encoding variational bayes,'' {\em arXiv
  preprint arXiv:1312.6114}  (2013).

\bibitem{mao2017least}
Mao, X., Li, Q., Xie, H., Lau, R.~Y., Wang, Z., and Smolley, S.~P., ``Least
  squares generative adversarial networks,'' in [{\em Computer Vision (ICCV),
  2017 IEEE International Conference on}{\nolinebreak\hspace{0.1em}]},
  2813--2821, IEEE (2017).

\bibitem{nair2010rectified}
Nair, V. and Hinton, G.~E., ``Rectified linear units improve restricted
  boltzmann machines,'' in [{\em Proceedings of the 27th international
  conference on machine learning (ICML-10)}{\nolinebreak\hspace{0.1em}]},
  807--814 (2010).

\bibitem{ulyanov1607instance}
Ulyanov, D., Vedaldi, A., and Lempitsky, V., ``Instance normalization: The
  missing ingredient for fast stylization. arxiv 2016,'' {\em arXiv preprint
  arXiv:1607.08022} .

\bibitem{ronneberger2015u}
Ronneberger, O., Fischer, P., and Brox, T., ``U-net: Convolutional networks for
  biomedical image segmentation,'' in [{\em International Conference on Medical
  image computing and computer-assisted
  intervention}{\nolinebreak\hspace{0.1em}]},   234--241, Springer (2015).

\bibitem{he2016deep}
He, K., Zhang, X., Ren, S., and Sun, J., ``Deep residual learning for image
  recognition,'' in [{\em Proceedings of the IEEE conference on computer vision
  and pattern recognition}{\nolinebreak\hspace{0.1em}]},   770--778 (2016).

\bibitem{eitz2012hdhso}
Eitz, M., Hays, J., and Alexa, M., ``How do humans sketch objects?,'' {\em ACM
  Trans. Graph. (Proc. SIGGRAPH)}~{\bf 31}(4),  44:1--44:10 (2012).

\end{thebibliography}

\end{document}